\documentclass{edm_article}

\usepackage{hyperref}
\hypersetup{
    colorlinks=true,
    linkcolor=blue,
    filecolor=magenta,      
    urlcolor=cyan,
    pdftitle={Overleaf Example},
    pdfpagemode=FullScreen,
    }

\usepackage{microtype}
\usepackage{array,ragged2e}
\usepackage{tabularray}
\usepackage{color, colortbl}
\usepackage{tikz}
\usetikzlibrary{positioning, shapes.geometric, arrows.meta}
\newcommand{\CC}[1]{\cellcolor[gray]{0.8}}

\begin{document}

\title{Investigating Automatic Scoring and Feedback using Large Language Models}


\numberofauthors{3}
\author{
    Gloria Ashiya Katuka\thanks{This work was completed during an internship at Coursera.}\\
    \affaddr{University of Florida}\\
    \email{gkatuka@ufl.edu}
    \and
    Alexander Gain\\
    \affaddr{Coursera}\\
    \email{again@coursera.org}
    \and
    Yen-Yun Yu\\
    \affaddr{Coursera}\\
    \email{yyu@coursera.org}
}

\maketitle

\begin{abstract}
Automatic grading and feedback have been long studied using traditional machine learning and deep learning techniques using language models. With the recent accessibility to high performing large language models (LLMs) like LLaMA-2, there is an opportunity to investigate the use of these LLMs for automatic grading and feedback generation. 

Despite the increase in performance, LLMs require significant computational resources for fine-tuning and additional specific adjustments to enhance their performance for such tasks. To address these issues, Parameter Efficient Fine-tuning (PEFT) methods, such as LoRA and QLoRA, have been adopted to decrease memory and computational requirements in model fine-tuning. This paper explores the efficacy of PEFT-based quantized models, employing classification or regression head, to fine-tune LLMs for automatically assigning continuous numerical grades to short answers and essays, as well as generating corresponding feedback. 

We conducted experiments on both proprietary and open-source datasets for our tasks. The results show that prediction of grade scores via finetuned LLMs are highly accurate, achieving less than 3\% error in grade percentage on average. For providing graded feedback fine-tuned 4-bit quantized LLaMA-2 13B models outperform competitive base models and achieve high similarity with subject matter expert feedback in terms of high BLEU and ROUGE scores and qualitatively in terms of feedback. 

The findings from this study provide important insights into the impacts of the emerging capabilities of using quantization approaches to fine-tune LLMs for various downstream tasks, such as automatic short answer scoring and feedback generation at comparatively lower costs and latency. 
\end{abstract}


\section{Introduction} 

With the rapid advancements of artificial intelligence (AI) and natural language processing (NLP) approaches, there has been increasing interest in developing more AI-powered grading and feedback systems for educational purposes \cite{lu2021integrating,jia2022automated, baral2021improving,suzen2020automatic, zhang2022automatic, qiu2022toward}. Research has extensively explored automatic scoring and feedback generation, yet natural language responses, specifically short answers and essays, present ongoing challenges due to their varying length, focus, and level of openness. \cite{burrows2015eras} With the advent and wider availability of language models, the potential for leveraging these technologies to enhance the capabilities of automatic grading and feedback systems has been a focus of investigation. These advancements are not intended to supplant educators but rather to augment their capabilities, offering time-saving benefits and enabling personalized feedback for students, a particularly valuable asset in settings where instructor resources are constrained, such as online education environments.

The introduction of high-capacity large language models (LLMs) like OpenAI's GPT \cite{radford2019language} and its successors, along with Meta's LLaMA-2 \cite{touvron2023llama2}, opens new avenues for exploring the application of cutting-edge LLMs in automated grading and feedback provision. While access to top-tier LLMs including GPT-3 \cite{brown2020language}, GPT-4 \cite{openai2023gpt4}, Chinchilla \cite{hoffmann2022training}, and both iterations of PaLM \cite{chowdhery2022palm, anil2023palm} remains restricted, the release of models like Falcon \cite{almazrouei2023falcon}, LLaMA \cite{touvron2023llama}, and LLaMA-2 \cite{touvron2023llama2} under open-access terms has democratized access to powerful LLMs. This newfound accessibility enables their fine-tuning for a range of specialized tasks, including the automation of grading and feedback generation.

Fine-tuning LLMs involve adapting the expansive knowledge of the pretrained LLMs for a target task, leading to the successful impacts of LLMs across many fields and applications \cite{wei2021finetuned}. In the context of automatic grading and feedback generation, fine-tuning these LLM often require two distinct approaches: For automatic grading, one common approach involves using discriminative techniques, such as classification and regression; whereas generative methods, which have grown in popularity due to the rise of GenAI, utilize generative models. With most high-performing LLMs primarily designed for generative tasks, our work aims to investigate the use of these LLMs for regression purposes as well, marking an innovative approach by integrating both capabilities into a unified system for advanced grading and feedback generation.

Despite their improved performance, a significant challenge with LLMs is their demand for considerable computational resources during fine-tuning and inference phases. To mitigate this, there has been a growing interest in quantization techniques \cite{xiao2023smoothquant, yao2023comprehensive}. Quantization is done by compressing floating-point numbers to lower bit width numbers like int8 and int4. These techniques aim to reduce memory and computational demands without substantially compromising the model's performance. Such strategies fall under the umbrella of Parameter-efficient fine-tuning (PEFT), as discussed in \cite{liu2022few}, where the focus is on fine-tuning a small number of parameters while retaining the core capabilities of LLMs. However, PEFT has mainly been applied for generative fine-tuning tasks as opposed to discriminative tasks like classification and regression. 

In this work, we explored the effectiveness of fine-tuned 4-bit quantized LLama-2 models for automatically grading grading and feedback generation on our proprietary dataset and an open source dataset \cite{filighera2022your}. We conducted several experiments to examine the possibilities for an LLM-based grading and feedback system using quantized models. Specifically, we made adjustment to the model architecture for a regression tasks and then utilized supervised instruction fine-tuning, a well-known approach for finetuning LLMs. In particular, we investigate the following research questions: 
\begin{itemize}
    \item \textbf{RQ1:} Can fine-tuning \textit{quantized} LLaMA-2 be leveraged to improve upon existing ML/DL approaches for automatic grading?
    \item \textbf{RQ2:} Can fine-tuning \textit{quantized} LLaMA-2 be leveraged to improve upon existing ML/DL approaches for automatic feedback generation?
    \item \textbf{RQ3:} Can combining the regression and generative approaches lead to higher quality feedback generation?
\end{itemize}

To this end, we conducted experiments on open-source and proprietary datasets for our tasks. We demonstrate that fine-tuned LLMs, including a 4-bit quantized version of the LLaMA-2 13B model, can predict grades with remarkable accuracy, averaging less than 3\% error in grade percentage. For providing graded feedback fine-tuned 4-bit quantized LLaMA-2 13B models outperform competitive base models and achieve high similarity with subject matter expert feedback in terms of high BLEU and ROUGE scores and qualitatively in terms of feedback. The findings from this study will provide important insights into the impacts of the emerging capabilities of using quantization approaches to fine-tune LLMs for various downstream tasks, such as automatic short answer scoring and feedback generation, offering a more cost-effective and efficient solution while maintaining high accuracy and quality of feedback.

\section{Background \& Related Work}
The emergence of transformer models in 2017 have revolutionalized NLP, with transformer models serving as state-of-the-art (SOTA) baselines for the many NLP tasks \cite{vaswani2017attention}. Pretrained transformer-based models (PTMs) such as Bert \cite{devlin2018bert}, RoBerTa \cite{liu2019roberta}, T5 \cite{raffel2020exploring}, GPT-2 \cite{radford2019language}, which have served as the foundation of advanced LLMs such as OpenAI's GPT-4  \cite{openai2023gpt4}, Google's PaLM-2 \cite{anil2023palm} and Meta's Llama-2 \cite{touvron2023llama} are becoming increasingly popular. The main difference between LLMs is the pretraining strategies utilized. Based on the transformer architecture, LLMs can be trained based on an encoder-only, decoder-only, or encoder-decoder transformer-based architectures \cite{vaswani2017attention}. Another important difference between LLMs is whether the model was trained for discriminative or generative purposes. Discriminative model training involves classifying or predicting an output from a set of categories, focusing on understanding and categorizing input data. This is different from generative models, which aim to generate new content. Discriminative models are commonly used in encoder-only and encoder-decoder based models, such as BERT \cite{devlin2018bert}, XLNet \cite{yang2019xlnet}, RoBERTa \cite{liu2019roberta}, and T5 \cite{raffel2020exploring}, mainly applying masked language modeling. Generative model training, on the other hand, has become a common approach for some of the most popular LLMs such as GPT-3 \cite{brown2020language}, GPT-4 \cite{openai2023gpt4}, PaLM \cite{chowdhery2022palm}, Llama \cite{touvron2023llama}, and Llama-2 \cite{touvron2023llama2}. These models are autoregressive in nature and have been trained for next-word prediction.

Pretraining LLMs provides an effective starting point for many NLP tasks; as they can be fine-tuned for specific tasks to yield better performance. Fine-tuning LLMs has been found to improve model performance and generalization \cite{xu2021raise}. Recently, researchers have explored supervised instruct-tuning, especially for newer LLMs using PEFT quantization approaches such as LoRA, QLoRA. 
PEFT is used to down-cast the data types from 32-bit float into lower precision data types. Down-casting data types to make model training faster is not a new idea. PERT with LoRA involves using the low-ranking matrices to recover the fine-tuned weight matrix and then added to the original model weight to get the final weights. \cite{dettmers2024qlora}

\section{Experiment settings} 
In this section we present our experiments for automatic scoring using regression models and feedback generation using the generative large language models. We experimented on two datasets: an open-source short answer and feedback dataset (English version)\cite{filighera2022your} \footnote{\href{https://huggingface.co/datasets/Short-Answer-Feedback/saf_communication_networks_english}{Link to dataset on HuggingFace }} and an proprietary short-answer and essay dataset.  Our goal was to test the hypotheses that the 4-bit quantized version of LLaMA-2 model will perform on-par or better than other LLMs for both regression and generative tasks, with the idea that combining the two approaches can lead to even better performance. We describe this approach is section 4. 

\subsection{Datasets}
\textbf{Short Answer and Feedback (SAF) dataset} \cite{filighera2022your}: The SAF dataset was recently introduced by Filighera et al. as a comprehensive dataset that can be used for both automatic grading and feedback generation. The original dataset presented in the paper is comprised of an assortment of both English and German short answer questions. For our study, we are using the version that contains 31 English-only questions covering a college level communications networks topics. The dataset was split into 1700 instances for training, 427 instances for validation, 375 instance for testing unseen answers to the questions in the train set and 479 instances of for testing unseen question that do not appear in the train set. Each student’s answer was scored by two graduate students, who had completed the communications networks course and two experienced appJobber employees using an annotation guide provided by the researcher. All English answers were annotated twice. 

\textbf{Proprietary dataset}: This is a proprietary dataset consisting of assessment questions, student answers, with accompanying graded scores and feedback for the answers, in a variety of subjects.  

\subsection{Choice of Pretrained Models}
The transformer model architecture, introduced by Vaswani et al. \cite{vaswani2017attention}, has emerged as a powerful paradigm for natural language processing tasks. Unlike traditional recurrent or convolutional architectures, transformers leverage self-attention mechanisms to capture global dependencies and contextual information efficiently. This capability makes transformers highly effective for various natural language discriminative and generative tasks. Pretrained Transformer Models (PTM) have significantly advanced the field of Natural Language Processing (NLP), thanks to the extensive knowledge derived from vast training data. For this work, we will be fine-tuning the following pretrained transformer models:

\subsubsection{\textbf{RoBERTa}} 
RoBERTa \cite{liu2019roberta}, Robustly Optimized BERT approach, builds on the original BERT and modifies the pretraining strategies, such as using byte-pair encoding \cite{shibata1999byte, bostrom2020byte}, modifying BERT's static MLM objective to a dynamic MLP, removing the next-sentence pretraining objectives and modifying key training parameters. Recently, RoBERTa has been found to outperform other traditional deep learning and BERT models for DA classification tasks \cite{duran2023sentence}.

\subsubsection{\textbf{GPT-2}} 
GPT-2 \cite{radford2019language} is a second generation variant of the GPT, proposed by Radford et al., focusing on generative language modeling. GPT models are decoder-only transformer-based models pretrained on large-scale datasets. GPT uses a causal language modeling objective and is therefore powerful at predicting the next token in a sequence. GPT-based models have been successful in tasks such as text classification, summarization, and question answering. The models leverage the autoregressive nature of the Transformer architecture to generate high-quality text samples.

\subsubsection{\textbf{LLaMA-2}}
LLaMA-2 \cite{touvron2023llama2} is a collection of newly released open-source LLMs based on the LLaMA \cite{touvron2023llama} by Meta GenAI. The release of these open-source LLaMA-2 models creates opportunities for the research community to fine-tune the actual weights and biases of the models with transparency and visibility to the model architecture and pretraining process. However, like most recent LLMs, LLaMA-2 is a decoder-only transformer model developed mainly for generative tasks.



  

\textbf{\textit{4-bit Quantization of LLaMA-2}}: Despite the open access to LLaMA-2 models, the high computational demands pose significant challenges. For instance, fine-tuning a LLaMA-2 7B model with full precision requires approximately 112GB of GPU memory, exceeding the capacity of consumer GPUs. To mitigate this, there has been a growing interest in parameter efficient fine-tuning (PEFT) \cite{houlsby2019parameter} quantization approaches. Recently, 4-bit quantization has shown optimal performance resulting in reduced latency and memory use \cite{dettmers2023case}. Equation \ref{quant}
shows the formula for quantizing a 32-bit Floating Point (FP32) tensor into a Int4 tensor with magnitude of [-7,7].

\vspace*{-2mm}
{
\small
\begin{equation}
    X_{\text{Int4}} = \text{round}\left(\frac{7}{\text{absmax}(X_{\text{FP32}})} \times X_{\text{FP32}}\right)
    \label{quant}
\end{equation}
}

\subsection{Implementation details}
In our experiments, all implementation was done in PyTorch \cite{paszke2019pytorch}. For each fine-tuning experiment with both variants of RoBERTa, we set the following hyperparameters: we used a batch size of 16 with an AdamW Optimizer with a learning rate of 1 e-5 and weight decay of 0.01. We trained for 20 epochs, with early stopping set at 10. For LLaMA-2 model variants, we use the \textit{bitandbytes} \cite{dettmers2022llm} library for the model quantization configuration. We attempted to use QLoRA \cite{dettmers2024qlora} with LoRA \cite{hu2021lora}, which enabled us to fine-tune only about 1\% of the parameters, but we faced the challenge of fine-tuning the LoRA and the 4bit model with the second corpus for our cross-corpora fine-tuning approaches, so we used the QLoRA configuration without LoRA, fine-tuning about 3.9\% of the parameters. We trained the quantized model for 10 epochs  using a batch size of 4 with a learning rate of 2 e-4 and a maximum sequence length of 512. To save memory, we use a paged 32-bit AdamW optimizer\cite{kingma2014adam} and weight decay of 0.05 and mixed precision \cite{micikevicius2017mixed}. All training was done using single NVIDIA A100 GPU. 

\subsection{Evaluation Metrics}
For the regression tasks, we evaluated the models based on their Root Mean Square Error (RMSE). Mean Absolute Error (MAE), and Pearson's correlation coefficient ($p$), providing a dual perspective on both the accuracy and correlation strength of predictions relative to the true values. 


The lower values of MAE and RMSE, and the higher values for $p$, BLEU, \& ROUGE correspond to bette rmodel outputs.

\section{Results and Discussion}
To evaluate the performance of the fine-tuned grading regression model, we compared the performance of the 4-bit quantized LLaMa-2 models with other existing open-source LLMs for the open-source and proprietary datasets. We then conduct experiments on graded feedback generation using similar methods and comparisons. 

\begin{figure*}[htb] 
\centering 

\begin{tikzpicture}[
    node distance=2cm and 0.5cm, 
    mynode/.style={draw, text width=3cm, align=center, rectangle, inner sep=5pt, minimum height=1.5cm}, 
    myarrow/.style={-Stealth},
    icon/.style={text width=1cm, node distance=0cm, inner sep=0pt},
]

\node[mynode] (qsa) {Question + Student Answer};
\node[mynode, right=of qsa] (gsdnn) {Grade Score Model};
\node[icon, right=of gsdnn] (cog1) {\includegraphics[width=1cm]{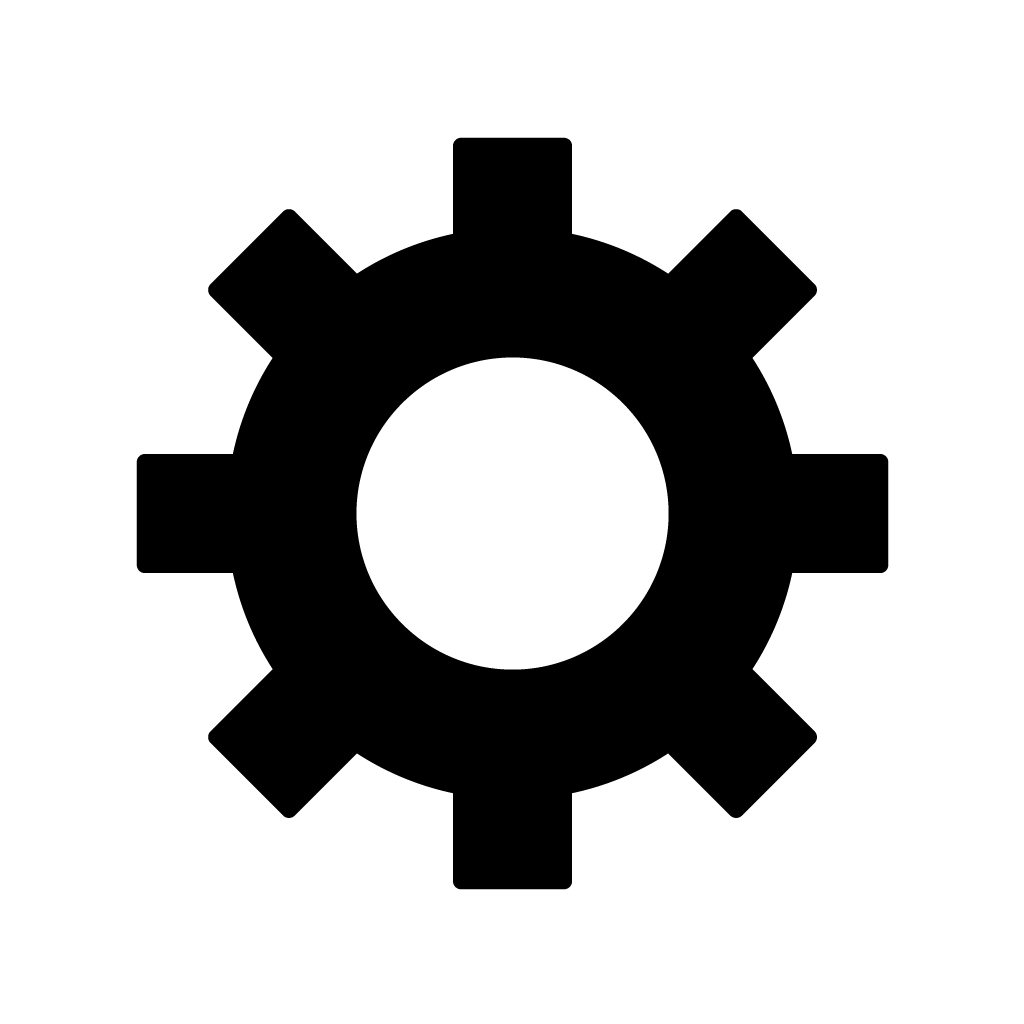}};
\node[mynode, right=of cog1] (llm) {Finetuned LLM};
\node[icon, right=of llm] (cog2) {\includegraphics[width=1cm]{./figures/cogwheel.png}};
\node[mynode, right=of cog2] (gaf) {Grade and feedback};
\node[mynode, above=of llm] (gr) {Grading rubric};

\draw[myarrow] (qsa) -- (gsdnn);
\draw[myarrow] (cog1) -- (llm);
\draw[myarrow] (cog2) -- (gaf);
\draw[myarrow] (gr) -- (llm);
\draw[myarrow] (qsa) |- ([yshift=-1.5cm]llm.south) -- (llm.south);

\end{tikzpicture}

\caption{An overview of our most performant models in terms of grading scores and grading feedback. Assessment questions, student answers, and grading rubrics are supplied as inputs to our grade score and feedback generation models. The grade score model outputs are supplied as input to our finetuned LLMs leading to higher performance. A highly accurate, w.r.t. domain expert ground-truth, grade score and feedback is outputted.}
\label{sys-diagram}
\end{figure*}

For feedback generation, comparison is done between LLMs that are supplied predicted grade scores versus those that are not. We show the former leads to better performance. Figure \ref{sys-diagram} shows an overview of our highest performing systems in terms of graded feedback and feedback generation. 

\subsection{Regression Models for Scores Prediction - SAF dataset} 
Similarly to the original paper, we compared the models with and without questions, Models with\_questions  received the questions as additional inputs, while models without\_questions did not. Table \ref{tab:saf_reg_results} reports the results of the regression models with fine-tuned RoBERTa, GPT-2 and LLaMa-2 7B evaluated on the unseen answers test set, and also highlights the results of the T5 model from the original paper.  

\begin{table}[htbp]
\caption{Comparison of scoring results on SAF dataset with former approaches
}
\label{tab:saf_reg_results}
\centering
\begin{tabular}{lccc}
\hline
\textbf{Model} & \textbf{RMSE} & \textbf{MAE} &  \textbf{\(\rho\)} \\
\hline
{\textbf{SAF dataset - Unseen Answers }} & & &\\
\hline
$T5_{wo\_quest}$ & 0.290 & - & -\\
$T5_{w\_quest}$  & 0.269 & - & -\\
\hline
$RoBERTa_{wo\_quest}$ & 0.268 & 0.151 & \textbf{0.939} \\
$GPT2_{wo\_quest}$ & 0.317 & 0.194 & 0.910 \\
$LLaMA-2 (7B)_{wo\_quest(lora)}  $ & 0.353 & 0.230 & 0.888 \\
$LLaMA-2 (7B)_{wo\_quest(qlora)}$  & 0.265 & 0.179 & 0.931 \\
$LLaMA-2 (7B)_{w\_quest(qlora)}$  & \textbf{0.257} & \textbf{0.168} & 0.934 \\
\hline
\end{tabular}

\end{table}

The results showed that the base RoBERTa model trained without questions in the input outperformed the benchmarked T5 model \cite{filighera2022your} with an RMSE of 0.268 and a $p$ value of 0.939. The quantized LLaMA-2 7B model outperformed the benchmarked T5 model with an RMSE of \textbf{0.257}. Interestingly, though the LLaMA-2 models were trained primarily for generative tasks, the model is able to outperform other models when fine-tuning is applied for regression tasks. 

\subsection{Regression Models for Scores Prediction - Proprietary dataset}
The experimental results presented in Table \ref{tab:c_reg_results} showcase the performance of various models on both the proprietary dataset and its upsampled version. In the context of the proprietary dataset, it is evident that the LLaMA-2-13B model with QLoRA consistently outperforms other models in terms of Root Mean Square Error (RMSE) and Mean Absolute Error (MAE), achieving an impressive RMSE of 0.036 and MAE of 0.028. Despite RoBERTa and GPT-2 displaying competitive results, with RoBERTa achieving an RMSE of 0.052 and GPT-2 achieving an RMSE of 0.050, their performance lags behind that of LLaMA-2-13B with QLoRA. However, it is noteworthy that RoBERTa and GPT-2 exhibit relatively weaker performance in terms of Spearman's rank correlation coefficient ($p$), with values of 0.423 and 0.115, respectively, compared to LLaMA-2-13B's $p$ of 0.512.

\begin{table}[htbp]
\caption{Experimental Results
}
\label{tab:c_reg_results}
\centering
\begin{tabular}{lccc}
\hline
\textbf{Model} & \textbf{RMSE} & \textbf{MAE} &   \textbf{\(\rho\)} \\

\hline
{\textbf{Proprietary dataset}} & & \\
\hline
RoBERTa & 0.052 & 0.039 & 0.423\\
GPT-2 & 0.050 & 0.036 & 0.115\\
LLaMA-2-7B w/QLoRA & 0.039 &  0.030 & 0.337 \\
LLaMA-2-13B w/QLoRA  & \textbf{0.036} & \textbf{0.028} &\textbf{0.512} \\
\hline
{\textbf{Proprietary \textit{upsampled}}} & & &\\
\hline
RoBERTa & 0.049 & 0.038 & 0.552\\
GPT-2 & 0.043 & 0.034 & 0.383\\
LLaMA-2-7B w/QLoRA & 0.040 & 0.030 & 0.401\\
LLaMA-2-13B w/QLoRA & \textbf{0.032} & \textbf{0.022} & \textbf{0.657} \\
\hline
\end{tabular}

\end{table}

Upon examining the results on the upsampled version of the proprietary dataset, we observe significant improvements across all models. Notably, LLaMA-2-13B with QLoRA achieves remarkable performance, yielding an RMSE of 0.032 and MAE of 0.022, which are notably lower compared to other models. This underscores the efficacy of the LLaMA-2-13B architecture combined with QLoRA in handling upsampled data. Additionally, the Spearman's rank correlation coefficient ($p$ ) for LLaMA-2-13B with QLoRA notably increases to 0.657, indicating its robustness in capturing the underlying relationships within the upsampled dataset. Overall, these results emphasize the importance of model architecture and data preprocessing techniques in enhancing the predictive performance of regression models, particularly in scenarios involving imbalanced datasets.

\subsection{Models for Feedback Generation}

For feedback generation, we compare the performance of our quantized LLaMA-2 models with LoRA to the base LLM model GPT-2. The outcomes of these experiments are valuable for two primary reasons: One is to test whether we are able to achieve high-similarity to expert domain ground-truth scores and feedback via the finetuning and quantization methods. Two is to test whether supplying of accurate grade score outputs is beneficial to performance. 

The outcomes of these experiments and the answers to the questions are seen in Table \ref{feedback-tab}. We finetune models for 20 epochs each with a constant learning rate of 2e-4, weight decay of 1e-3, and AdamW optimizer. The same 4-bit quantization and lora settings prior described are used.

\begin{figure}[ht]
    \centering
    \includegraphics[width=0.7\linewidth]{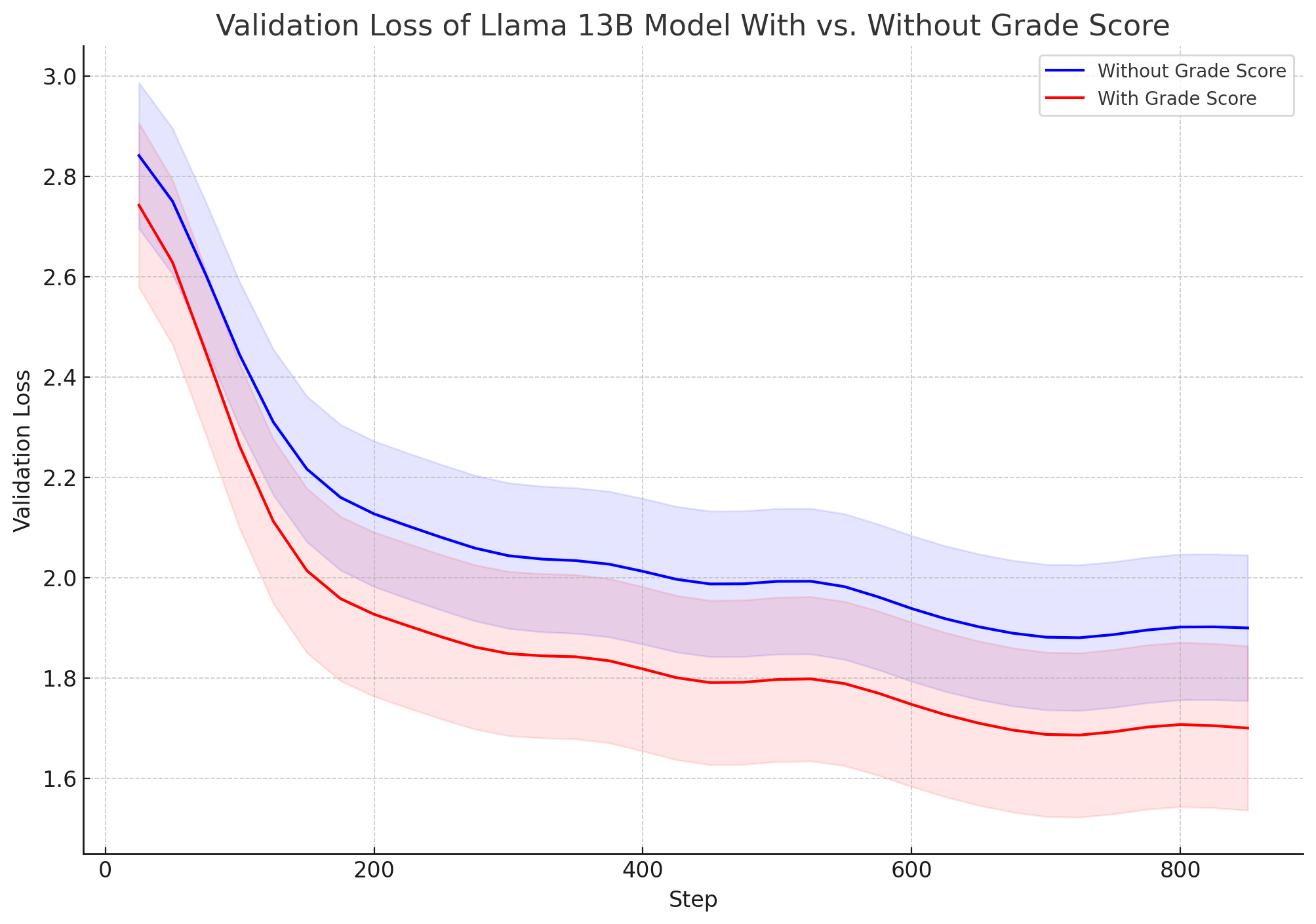}
    \caption{The plot shows the validation loss of the Llama 13B model over a subset of the training steps, comparing the performance with and without the application of a grade score supplied by the regression model. This plot highlights the impact of the grade score leading to better generalization and lower validation loss over time.}
    \label{fig:llama_validation_loss}
\end{figure}

We see that LLaMA-13B are most performnant with respect to expert feedback similarity as reported by relatively higher BLEU and ROUGE scores for both the SAR dataset and the proprietary dataset. The \texttt{w/ grade} rows of the table correspons to models where the outputted predicted score is supplied to the models as additional input. It's evident that supplying the grade scores provides increase in performance with LLaMA-13 variant with grade score supplied achieving the best results with BLEU, ROUGE-1, and ROUGE-2 scores of \textbf{0.396}, \textbf{0.232}, and \textbf{0.137} respectively for the SAR dataset and scores of \textbf{0.707}, \textbf{0.775}, and \textbf{0.737} for the proprietary dataset. This is further emphasized in Figure \ref{fig:llama_validation_loss} where there is clear advantage in terms of the validation loss and generalization error when supplying the grade score.

To provide qualitative assessment of some questions and answers from the SAR dataset, we show two randomly selected question, answer pairs  in Tables \ref{tab:examples_models} and \ref{tab:examples_models_congestion_control_fixed_with_background}. We show the question, provided answer, and ground truth score and feedback for each. We also show the predicted scores and feedback for the highest and lowest performing models, the LLaMA-2 model finetuned model with grade score supplied and the GPT-2 model without grade score supplied for contrast in terms of closeness to the ground-truth score, and quality of graded feedback. In terms of quality, the example in Table \ref{tab:examples_models} shows how the LLaMA-2 model has learned to give feedback that is similar in both style and content to the expert feedback. The example in Table \ref{tab:examples_models_congestion_control_fixed_with_background} showcases relatively more how the LLaMA-2 model is closer in technical content than the GPT-2 model with respect to the expert feedback. 

\begin{table}[htbp]
\caption{Experimental results for feedback generation on the SAR dataset}
\label{tab:sar_gen_results}
\centering
\begin{tabular}{lccc}
\hline
\textbf{Model} & \textbf{BLEU} & \textbf{ROUGE-1} &  \textbf{ROUGE-2} \\
\hline
{\textbf{SAR dataset}} & & &\\
\hline

GPT-2 & 0.17 & 0.025 & 0.012 \\
GPT-2 w/ grade & 0.206 & 0.058 & 0.017 \\
LLaMA-7B & 0.24 & 0.060 & 0.026 \\
LLaMA-7B w/ grade & 0.352 & 0.187 & 0.099 \\
LLaMA-13B & 0.31 & 0.12 & 0.053 \\
LLaMA-13B w/ grade & \textbf{0.396} & \textbf{0.232} & \textbf{0.137} \\

\hline

{\textbf{Proprietary dataset}} & & &\\
\hline

GPT-2 & 0.0218 & 0.292 & 0.062 \\
GPT-2 w/ grade & 0.061 & 0.315 & 0.085 \\
LLaMA-2-7B & 0.617 & 0.716 & 0.642 \\
LLaMA-2-7B w/ grade & 0.657 & 0.759 & 0.682 \\
LLaMA-2-13B & 0.667 & 0.732 & 0.697 \\
LLaMA-2-13B w/ grade & \textbf{0.707} & \textbf{0.775} & \textbf{0.737} \\

\hline
\end{tabular}
\label{feedback-tab}
\end{table}

\begin{table*}[htbp]
    \centering
    \begin{tabular}{m{0.25\textwidth}|m{0.70\textwidth}}
    \hline
    \CC{} \textbf{Question} & State at least 4 of the differences shown in the lecture between the UDP and TCP headers. \\
    \hline
    \CC{} \textbf{Provided\_Answer} & While both (UDP and TCP headers) have source and destination port fields, every other part of their headers differ. For UDP, only a packet length and a checksum field follows the two previously mentioned fields. TCP needs more information. So after the source and destination port, the header is followed by a sequence number field as well as a field for the ack number. The TCP header also stores information on HL/RESV/Flags, Window size, checksum (as in UDP), Urgent Pointer, and Options. \\
    \hline  
    \CC{}\textbf{Ground Truth Score \& \newline Feedback}  & 1.0 \newline The response correctly identifies four differences between TCP and UDP headers. \\
    \hline  
    \multicolumn{2}{c}{\CC{}\textbf{LLaMA-2 Model}} \\
    \hline
    \CC{} \textbf{Predicted Score \& \newline Feedback} & 1.0 \newline The response correctly identifies four differences between TCP and UDP headers. full marks will be awarded for this. \bigskip\bigskip\\
    \hline 
    \multicolumn{2}{c}{\CC{}\textbf{GPT-2 Model}} \\
    \hline
    \CC{} \textbf{Predicted Score \& \newline Feedback} & 1.0 \newline It correctly answers the request correctly. \bigskip\bigskip\\
    \hline 
    \end{tabular}
    \caption{Example inputs and outputs of the LLAMA-2 and GPT-2 models}.
    \label{tab:examples_models}
\end{table*}

\begin{table*}[htbp]
    \centering
    \begin{tabular}{m{0.25\textwidth}|m{0.70\textwidth}}
    \hline
    \CC{} \textbf{Question} & In the lecture you have learned about congestion control with TCP. Name the 2 phases of congestion control and explain how the Congestion Window (\texttt{cwnd}) and the Slow Start Threshold (\texttt{ss\_thresh}) change in each phase (after initialization, where \texttt{cwnd} = 1 and \texttt{ss\_thresh} = advertised window size) in 1-4 sentences. \\
    \hline
    \CC{} \textbf{Provided\_Answer} & The first phase is called ``slow start''. After the initialisation, the sender starts sending segments, and waits for the receiver to acknowledge them all. This number will double every Round Trip Time (RTT) until the advertised window size is reached. If a timeout happens beforehand, phase one is restarted immediately. \newline When \texttt{ss\_thresh} is reached, phase two - ``congestion avoidance'' - is entered, when the RTT will be increased linearly until a timeout occurs. When this occurs, phase one is initialized again. These two phases will be repeated over and over again, the sending rate will never be constant with TCP. \\
    \hline  
    \CC{} \textbf{Ground Truth Score \& Feedback}  & 0.75 \newline The response is correct except that in both the phases when the congestion occurs, \texttt{ss\_thresh} is set to half of the current size of the congestion window and the congestion window is reset to one, which is not the same as the initialized value where \texttt{cwnd} = 1 and \texttt{ss\_thresh} = advertised window. \\
    \hline  
    \multicolumn{2}{c}{\CC{}\textbf{LLaMA-2 Model}} \\
    \hline
    \CC{} \textbf{Predicted Score \& Feedback} & 0.65 \newline The response correctly explains the slow start and congestion avoidance phases of TCP. However, the slow start phase is not entered immediately after the initialisation. Instead, the sender waits for the acknowledgement of the first segment before entering the slow start phase. In the congestion avoidance phase, the sender will decrease the congestion window by 50\% of the current window size. \bigskip\bigskip\\
    \hline 
    \multicolumn{2}{c}{\CC{}\textbf{GPT-2 Model}} \\
    \hline
    \CC{} \textbf{Predicted Score \& Feedback} & 0.50 \newline The response incorrectly describes a latency factor of 6. On the other hand, \texttt{ss\_thresh} uses 2x the latency factor with TCP only because its TCP latency factor is higher. The error incorrectly refers to slow start (RTT) as a time lag. The response correctly identifies different stages of slower start with TCP. The timings are not updated \bigskip\bigskip\\
    \hline 
    \end{tabular}
    \caption{Example inputs and outputs of the LLAMA-2 and GPT-2 models.}
    \label{tab:examples_models_congestion_control_fixed_with_background}
\end{table*}

\subsection{Implications}
The performance of the quantized LLaMA-2 models suggests that we can augment the automatic grading and feedback process in educational technologies, thereby significantly enhancing the efficiency and scalability of online learning environments. These advancements are intended to augment instructor capabilities, offering time-saving benefits and enabling personalized feedback for students, a particularly valuable asset in settings where instructor resources are constrained, such as online education environments.

\section{Conclusion and Future work}

In this study, we have presented preliminary work towards developing an auto-scoring method for student response and feedback generation using quantized LLaMA-2 models. The quantized LLaMA-2 13B model, fine-tuned with QLoRA, has shown exceptional performance in terms of accuracy, achieving a remarkable reduction in error rates and outperforming baseline models in grading short answers and essays. 

Furthermore, the application of these models to feedback generation has yielded promising results, with quantized LLaMA-2 models surpassing traditional LLMs like GPT-2 in generating feedback that closely aligns with expert evaluations. The incorporation of predicted grade scores as additional input further enhanced the model's performance.

Our results offer significant insights into the implications of employing quantization techniques for fine-tuning large language models (LLMs) for a variety of downstream applications, including automatic scoring of short answers and generating feedback, at reduced costs and lower latency.

\section*{Limitations}

Due to computing limitations, we were unable to investigate the scaling behavior of the LLMs, such as investigating with different precision and with larger models. Further experiments and studies are need in the future to investigate the impact of fine-tuning significantly larger LLMs, and whether such LLMs can still be deployed cost-effectively.


%
\bibliographystyle{abbrv}
\bibliography{sigproc}  

\begin{thebibliography}{10}

\bibitem{almazrouei2023falcon}
E.~Almazrouei, H.~Alobeidli, A.~Alshamsi, A.~Cappelli, R.~Cojocaru, M.~Debbah, E.~Goffinet, D.~Heslow, J.~Launay, Q.~Malartic, et~al.
\newblock Falcon-40b: an open large language model with state-of-the-art performance.
\newblock Technical report, Technical report, Technical report, Technology Innovation Institute, 2023.

\bibitem{anil2023palm}
R.~Anil, A.~M. Dai, O.~Firat, M.~Johnson, D.~Lepikhin, A.~Passos, S.~Shakeri, E.~Taropa, P.~Bailey, Z.~Chen, E.~Chu, J.~H. Clark, L.~E. Shafey, Y.~Huang, K.~Meier-Hellstern, G.~Mishra, E.~Moreira, M.~Omernick, K.~Robinson, S.~Ruder, Y.~Tay, K.~Xiao, Y.~Xu, Y.~Zhang, G.~H. Abrego, J.~Ahn, J.~Austin, P.~Barham, J.~Botha, J.~Bradbury, S.~Brahma, K.~Brooks, M.~Catasta, Y.~Cheng, C.~Cherry, C.~A. Choquette-Choo, A.~Chowdhery, C.~Crepy, S.~Dave, M.~Dehghani, S.~Dev, J.~Devlin, M.~Díaz, N.~Du, E.~Dyer, V.~Feinberg, F.~Feng, V.~Fienber, M.~Freitag, X.~Garcia, S.~Gehrmann, L.~Gonzalez, G.~Gur-Ari, S.~Hand, H.~Hashemi, L.~Hou, J.~Howland, A.~Hu, J.~Hui, J.~Hurwitz, M.~Isard, A.~Ittycheriah, M.~Jagielski, W.~Jia, K.~Kenealy, M.~Krikun, S.~Kudugunta, C.~Lan, K.~Lee, B.~Lee, E.~Li, M.~Li, W.~Li, Y.~Li, J.~Li, H.~Lim, H.~Lin, Z.~Liu, F.~Liu, M.~Maggioni, A.~Mahendru, J.~Maynez, V.~Misra, M.~Moussalem, Z.~Nado, J.~Nham, E.~Ni, A.~Nystrom, A.~Parrish, M.~Pellat, M.~Polacek, A.~Polozov, R.~Pope, S.~Qiao, E.~Reif, B.~Richter,
  P.~Riley, A.~C. Ros, A.~Roy, B.~Saeta, R.~Samuel, R.~Shelby, A.~Slone, D.~Smilkov, D.~R. So, D.~Sohn, S.~Tokumine, D.~Valter, V.~Vasudevan, K.~Vodrahalli, X.~Wang, P.~Wang, Z.~Wang, T.~Wang, J.~Wieting, Y.~Wu, K.~Xu, Y.~Xu, L.~Xue, P.~Yin, J.~Yu, Q.~Zhang, S.~Zheng, C.~Zheng, W.~Zhou, D.~Zhou, S.~Petrov, and Y.~Wu.
\newblock Palm 2 technical report, 2023.

\bibitem{baral2021improving}
S.~Baral, A.~F. Botelho, J.~A. Erickson, P.~Benachamardi, and N.~T. Heffernan.
\newblock Improving automated scoring of student open responses in mathematics.
\newblock {\em International Educational Data Mining Society}, 2021.

\bibitem{bostrom2020byte}
K.~Bostrom and G.~Durrett.
\newblock Byte pair encoding is suboptimal for language model pretraining.
\newblock {\em arXiv preprint arXiv:2004.03720}, 2020.

\bibitem{brown2020language}
T.~Brown, B.~Mann, N.~Ryder, M.~Subbiah, J.~D. Kaplan, P.~Dhariwal, A.~Neelakantan, P.~Shyam, G.~Sastry, A.~Askell, et~al.
\newblock Language models are few-shot learners.
\newblock {\em Advances in neural information processing systems}, 33:1877--1901, 2020.

\bibitem{burrows2015eras}
S.~Burrows, I.~Gurevych, and B.~Stein.
\newblock The eras and trends of automatic short answer grading.
\newblock {\em International journal of artificial intelligence in education}, 25:60--117, 2015.

\bibitem{chowdhery2022palm}
A.~Chowdhery, S.~Narang, J.~Devlin, M.~Bosma, G.~Mishra, A.~Roberts, P.~Barham, H.~W. Chung, C.~Sutton, S.~Gehrmann, et~al.
\newblock Palm: Scaling language modeling with pathways.
\newblock {\em arXiv preprint arXiv:2204.02311}, 2022.

\bibitem{dettmers2022llm}
T.~Dettmers, M.~Lewis, Y.~Belkada, and L.~Zettlemoyer.
\newblock Llm. int8 (): 8-bit matrix multiplication for transformers at scale.
\newblock {\em arXiv preprint arXiv:2208.07339}, 2022.

\bibitem{dettmers2024qlora}
T.~Dettmers, A.~Pagnoni, A.~Holtzman, and L.~Zettlemoyer.
\newblock Qlora: Efficient finetuning of quantized llms.
\newblock {\em Advances in Neural Information Processing Systems}, 36, 2024.

\bibitem{dettmers2023case}
T.~Dettmers and L.~Zettlemoyer.
\newblock The case for 4-bit precision: k-bit inference scaling laws.
\newblock In {\em International Conference on Machine Learning}, pages 7750--7774. PMLR, 2023.

\bibitem{devlin2018bert}
J.~Devlin, M.-W. Chang, K.~Lee, and K.~Toutanova.
\newblock Bert: Pre-training of deep bidirectional transformers for language understanding.
\newblock {\em arXiv preprint arXiv:1810.04805}, 2018.

\bibitem{duran2023sentence}
N.~Duran, S.~Battle, and J.~Smith.
\newblock Sentence encoding for dialogue act classification.
\newblock {\em Natural Language Engineering}, 29(3):794--823, 2023.

\bibitem{filighera2022your}
A.~Filighera, S.~Parihar, T.~Steuer, T.~Meuser, and S.~Ochs.
\newblock Your answer is incorrect... would you like to know why? introducing a bilingual short answer feedback dataset.
\newblock In {\em Proceedings of the 60th Annual Meeting of the Association for Computational Linguistics (Volume 1: Long Papers)}, pages 8577--8591, 2022.

\bibitem{hoffmann2022training}
J.~Hoffmann, S.~Borgeaud, A.~Mensch, E.~Buchatskaya, T.~Cai, E.~Rutherford, D.~d.~L. Casas, L.~A. Hendricks, J.~Welbl, A.~Clark, et~al.
\newblock Training compute-optimal large language models.
\newblock {\em arXiv preprint arXiv:2203.15556}, 2022.

\bibitem{houlsby2019parameter}
N.~Houlsby, A.~Giurgiu, S.~Jastrzebski, B.~Morrone, Q.~De~Laroussilhe, A.~Gesmundo, M.~Attariyan, and S.~Gelly.
\newblock Parameter-efficient transfer learning for nlp.
\newblock In {\em International Conference on Machine Learning}, pages 2790--2799. PMLR, 2019.

\bibitem{hu2021lora}
E.~J. Hu, Y.~Shen, P.~Wallis, Z.~Allen-Zhu, Y.~Li, S.~Wang, L.~Wang, and W.~Chen.
\newblock Lora: Low-rank adaptation of large language models.
\newblock {\em arXiv preprint arXiv:2106.09685}, 2021.

\bibitem{jia2022automated}
Q.~Jia, M.~Young, Y.~Xiao, J.~Cui, C.~Liu, P.~Rashid, E.~Gehringer, et~al.
\newblock Automated feedback generation for student project reports: A data-driven approach.
\newblock {\em Journal of Educational Data Mining}, 14(3):132--161, 2022.

\bibitem{kingma2014adam}
D.~P. Kingma and J.~Ba.
\newblock Adam: A method for stochastic optimization.
\newblock {\em arXiv preprint arXiv:1412.6980}, 2014.

\bibitem{liu2022few}
H.~Liu, D.~Tam, M.~Muqeeth, J.~Mohta, T.~Huang, M.~Bansal, and C.~A. Raffel.
\newblock Few-shot parameter-efficient fine-tuning is better and cheaper than in-context learning.
\newblock {\em Advances in Neural Information Processing Systems}, 35:1950--1965, 2022.

\bibitem{liu2019roberta}
Y.~Liu, M.~Ott, N.~Goyal, J.~Du, M.~Joshi, D.~Chen, O.~Levy, M.~Lewis, L.~Zettlemoyer, and V.~Stoyanov.
\newblock Roberta: A robustly optimized bert pretraining approach.
\newblock {\em arXiv preprint arXiv:1907.11692}, 2019.

\bibitem{lu2021integrating}
C.~Lu and M.~Cutumisu.
\newblock Integrating deep learning into an automated feedback generation system for automated essay scoring.
\newblock {\em International Educational Data Mining Society}, 2021.

\bibitem{micikevicius2017mixed}
P.~Micikevicius, S.~Narang, J.~Alben, G.~Diamos, E.~Elsen, D.~Garcia, B.~Ginsburg, M.~Houston, O.~Kuchaiev, G.~Venkatesh, et~al.
\newblock Mixed precision training.
\newblock {\em arXiv preprint arXiv:1710.03740}, 2017.

\bibitem{openai2023gpt4}
OpenAI.
\newblock Gpt-4 technical report, 2023.

\bibitem{paszke2019pytorch}
A.~Paszke, S.~Gross, F.~Massa, A.~Lerer, J.~Bradbury, G.~Chanan, T.~Killeen, Z.~Lin, N.~Gimelshein, L.~Antiga, et~al.
\newblock Pytorch: An imperative style, high-performance deep learning library.
\newblock {\em Advances in neural information processing systems}, 32, 2019.

\bibitem{qiu2022toward}
W.~Qiu, S.~Supraja, and A.~W. Khong.
\newblock Toward better grade prediction via a2gp--an academic achievement inspired predictive model.
\newblock {\em International Educational Data Mining Society}, 2022.

\bibitem{radford2019language}
A.~Radford, J.~Wu, R.~Child, D.~Luan, D.~Amodei, I.~Sutskever, et~al.
\newblock Language models are unsupervised multitask learners.
\newblock {\em OpenAI blog}, 1(8):9, 2019.

\bibitem{raffel2020exploring}
C.~Raffel, N.~Shazeer, A.~Roberts, K.~Lee, S.~Narang, M.~Matena, Y.~Zhou, W.~Li, and P.~J. Liu.
\newblock Exploring the limits of transfer learning with a unified text-to-text transformer.
\newblock {\em The Journal of Machine Learning Research}, 21(1):5485--5551, 2020.

\bibitem{shibata1999byte}
Y.~Shibata, T.~Kida, S.~Fukamachi, M.~Takeda, A.~Shinohara, T.~Shinohara, and S.~Arikawa.
\newblock Byte pair encoding: A text compression scheme that accelerates pattern matching.
\newblock 1999.

\bibitem{suzen2020automatic}
N.~S{\"u}zen, A.~N. Gorban, J.~Levesley, and E.~M. Mirkes.
\newblock Automatic short answer grading and feedback using text mining methods.
\newblock {\em Procedia computer science}, 169:726--743, 2020.

\bibitem{touvron2023llama}
H.~Touvron, T.~Lavril, G.~Izacard, X.~Martinet, M.-A. Lachaux, T.~Lacroix, B.~Rozi{\`e}re, N.~Goyal, E.~Hambro, F.~Azhar, et~al.
\newblock Llama: Open and efficient foundation language models.
\newblock {\em arXiv preprint arXiv:2302.13971}, 2023.

\bibitem{touvron2023llama2}
H.~Touvron, L.~Martin, K.~Stone, P.~Albert, A.~Almahairi, Y.~Babaei, N.~Bashlykov, S.~Batra, P.~Bhargava, S.~Bhosale, D.~Bikel, L.~Blecher, C.~C. Ferrer, M.~Chen, G.~Cucurull, D.~Esiobu, J.~Fernandes, J.~Fu, W.~Fu, B.~Fuller, C.~Gao, V.~Goswami, N.~Goyal, A.~Hartshorn, S.~Hosseini, R.~Hou, H.~Inan, M.~Kardas, V.~Kerkez, M.~Khabsa, I.~Kloumann, A.~Korenev, P.~S. Koura, M.-A. Lachaux, T.~Lavril, J.~Lee, D.~Liskovich, Y.~Lu, Y.~Mao, X.~Martinet, T.~Mihaylov, P.~Mishra, I.~Molybog, Y.~Nie, A.~Poulton, J.~Reizenstein, R.~Rungta, K.~Saladi, A.~Schelten, R.~Silva, E.~M. Smith, R.~Subramanian, X.~E. Tan, B.~Tang, R.~Taylor, A.~Williams, J.~X. Kuan, P.~Xu, Z.~Yan, I.~Zarov, Y.~Zhang, A.~Fan, M.~Kambadur, S.~Narang, A.~Rodriguez, R.~Stojnic, S.~Edunov, and T.~Scialom.
\newblock Llama 2: Open foundation and fine-tuned chat models, 2023.

\bibitem{vaswani2017attention}
A.~Vaswani, N.~Shazeer, N.~Parmar, J.~Uszkoreit, L.~Jones, A.~N. Gomez, {\L}.~Kaiser, and I.~Polosukhin.
\newblock Attention is all you need.
\newblock {\em Advances in neural information processing systems}, 30, 2017.

\bibitem{wei2021finetuned}
J.~Wei, M.~Bosma, V.~Y. Zhao, K.~Guu, A.~W. Yu, B.~Lester, N.~Du, A.~M. Dai, and Q.~V. Le.
\newblock Finetuned language models are zero-shot learners.
\newblock {\em arXiv preprint arXiv:2109.01652}, 2021.

\bibitem{xiao2023smoothquant}
G.~Xiao, J.~Lin, M.~Seznec, H.~Wu, J.~Demouth, and S.~Han.
\newblock Smoothquant: Accurate and efficient post-training quantization for large language models.
\newblock In {\em International Conference on Machine Learning}, pages 38087--38099. PMLR, 2023.

\bibitem{xu2021raise}
R.~Xu, F.~Luo, Z.~Zhang, C.~Tan, B.~Chang, S.~Huang, and F.~Huang.
\newblock Raise a child in large language model: Towards effective and generalizable fine-tuning.
\newblock {\em arXiv preprint arXiv:2109.05687}, 2021.

\bibitem{yang2019xlnet}
Z.~Yang, Z.~Dai, Y.~Yang, J.~Carbonell, R.~R. Salakhutdinov, and Q.~V. Le.
\newblock Xlnet: Generalized autoregressive pretraining for language understanding.
\newblock {\em Advances in neural information processing systems}, 32, 2019.

\bibitem{yao2023comprehensive}
Z.~Yao, C.~Li, X.~Wu, S.~Youn, and Y.~He.
\newblock A comprehensive study on post-training quantization for large language models.
\newblock {\em arXiv preprint arXiv:2303.08302}, 2023.

\bibitem{zhang2022automatic}
M.~Zhang, S.~Baral, N.~Heffernan, and A.~Lan.
\newblock Automatic short math answer grading via in-context meta-learning.
\newblock {\em arXiv preprint arXiv:2205.15219}, 2022.

\end{thebibliography}

\balancecolumns
\end{document}